%% file: neurips_2020.tex
\documentclass{article}

\usepackage[preprint, nonatbib]{neurips_2020}

\usepackage[utf8]{inputenc} 
\usepackage[T1]{fontenc}    
\usepackage{hyperref}       
\usepackage{url}            
\usepackage{booktabs}       
\usepackage{amsfonts}       
\usepackage{amssymb}
\usepackage{bm}
\usepackage{nicefrac}       
\usepackage{microtype}      
\usepackage{amsmath}
\usepackage{xcolor}
\usepackage{graphicx} 
\usepackage{adjustbox}
\usepackage{multirow}
\usepackage{enumitem}
\usepackage[normalem]{ulem}
\usepackage[numbers]{natbib}
\usepackage[skip=2pt,font=footnotesize,labelfont=bf]{caption}

\definecolor{our_blue}{rgb}{0.21747533000128158, 0.5305292836088684, 0.7548225041650647}
\definecolor{our_red}{rgb}{0.7364705882352941, 0.08, 0.10117647058823528}

\title{Improving Relation Extraction by Leveraging Knowledge Graph Link Prediction}

\author{
\textbf{George Stoica} \\
  Carnegie Mellon University\\
  5000 Forbes Ave, \\
  Pittsburgh, PA 15213 \\
  \texttt{gis@cs.cmu.edu} \\
   \And
  \textbf{Emmanouil Antonios Platanios} \\
  Microsoft Semantic Machines\\
  1 Microsoft Way, \\
  Redmond, WA 98052 \\
  \texttt{emplata@microsoft.com} \\
  \And
  \textbf{Barnabás Póczos} \\
  Carnegie Mellon University\\
  5000 Forbes Ave, \\
  Pittsburgh, PA 15213 \\
  \texttt{bapoczos@cs.cmu.edu} \\
}
\begin{document}

\maketitle

\begin{abstract}

Relation extraction (RE) aims to predict a relation between a subject and an object in a sentence, while knowledge graph link prediction (KGLP) aims to predict a set of objects, $O$, given a subject and a relation from a knowledge graph.
These two problems are closely related as their respective objectives are intertwined: given a sentence containing a subject and an object $o$, a RE model predicts a relation that can then be used by a KGLP model together with the subject, to predict a set of objects $O$.
Thus, we expect object $o$ to be in set $O$.
In this paper, we leverage this insight by proposing a multi-task learning approach that improves the performance of RE models by jointly training on RE and KGLP tasks.
We illustrate the generality of our approach by applying it on several existing RE models and empirically demonstrate how it helps them achieve consistent performance gains.

\end{abstract}

\section{Introduction}
\label{sec:intro}
\input{introduction}

\section{Background}
\label{sec:background}
\input{background}

\section{Proposed Method}
\label{sec:method}
\input{method}
\section{Experiments}
\label{sec:experiments}
\input{experiments}

\section{Related Work}
\label{sec:related_work}
\input{related_work}

\section{Conclusion}
\label{sec:conclusion}
\input{conclusion}


\bibliographystyle{plainnat}
\nocite{*}
\bibliography{ref}

\end{document}

%% file: introduction.tex
Many real-world applications ranging from search engines to conversational agents rely on the ability to uncover new relationships from existing knowledge.
Relation extraction (RE) and knowledge graph (KG) link prediction (KGLP) are two closely related tasks that center around inferring new information from existing facts. 
RE is the task of uncovering the relationship between two entities (termed the subject and object respectively) in a sentence. 
Similarly, KGLP involves inferring the set of correct answers (i.e., objects) to KG questions consisting of an entity (subject) and relation. 
These questions are given in triple-form: (\texttt{SUBJECT}, \texttt{RELATION}, ?).
To illustrate their relationship, consider the sentence ``\texttt{\textcolor{our_blue}{John} and \textcolor{our_red}{Jane} are married}'', whose subject and object are highlighted in \textcolor{our_blue}{blue} and \textcolor{our_red}{red} respectively.
Given this information, RE models infer the relationship between ``\texttt{\textcolor{our_blue}{John}}'' and ``\textcolor{our_red}{Jane}'' (e.g., ``\texttt{Spouse}'').
Similarly, KGLP models infer the answers (objects) to the question (\texttt{\textcolor{our_blue}{John}}, \texttt{Spouse}, ?).
Based on the sentence, the answers must include ``\texttt{\textcolor{our_red}{Jane}}''.
Thus, RE models predict the relation between a subject and object, while KGLP models infer the object from the subject and relation.

Several methods have been proposed to boost the performance of RE models by incorporating information from KGLP.
However, these approaches typically require KGLP pre-training \cite{weston-2013, lfds}, exhibit constrained parameter sharing \cite{weston-2013, lfds}, or predominately attend over both problems through custom attention mechanisms \cite{bag_re_kglp,han,long_tail}.
Moreover, these frameworks only support a limited class of KGLP models that can be reframed as inferring relations from subject and objects.
This constraint excludes recent KGLP methods which perform significantly better, but cannot be reformulated to satisfy the restriction.
An ideal framework should support arbitrary RE and KGLP methods, including the significantly more expressive and stronger performing recent KGLP approaches.
Additionally, such a framework should enable RE models to benefit from KGLP models with minimal changes to the underlying RE and KGLP methods.

\begin{figure}[t]
    \centering
    \includegraphics[width=.95\textwidth]{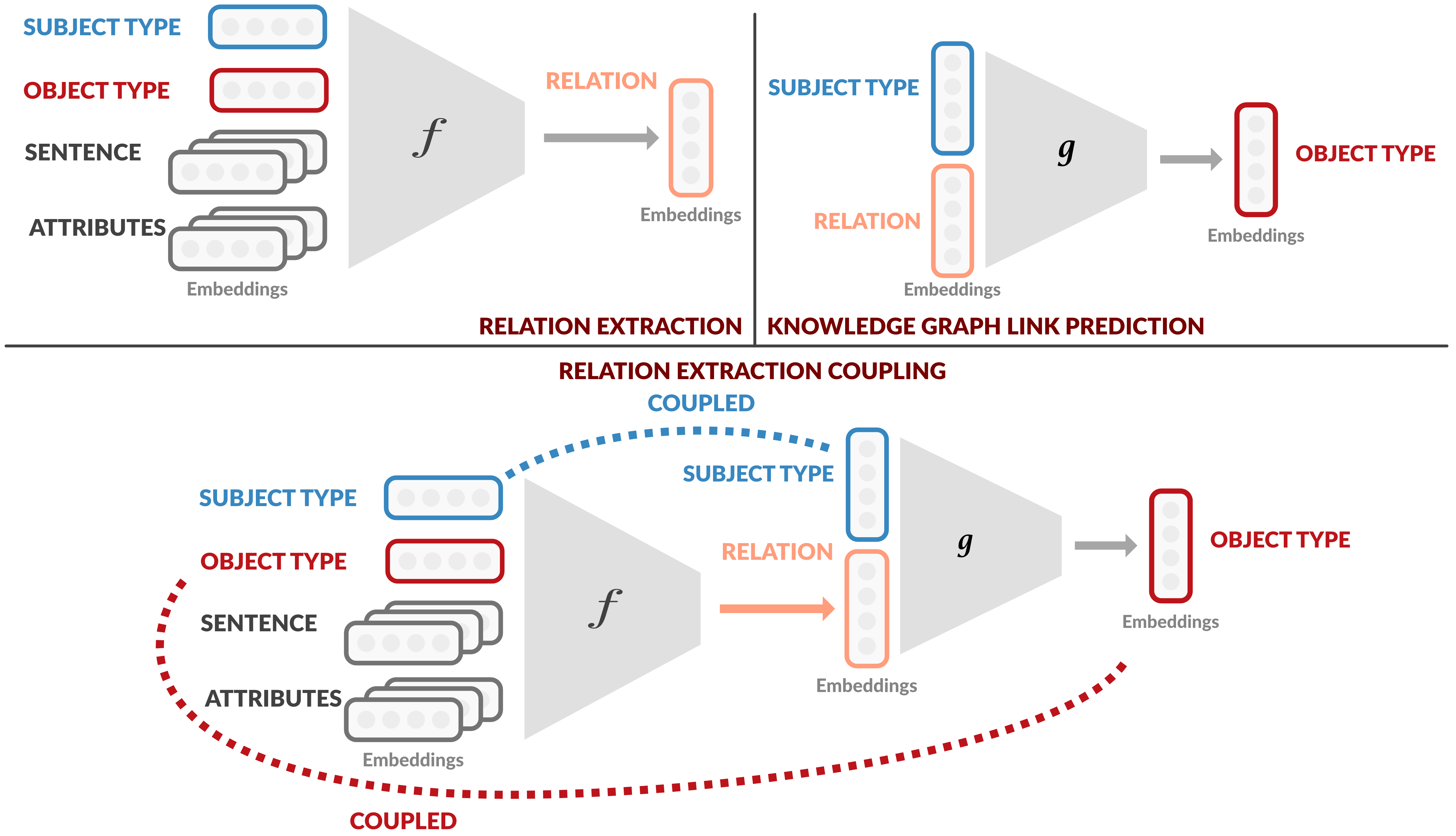}
    \caption{
        Overview of JRRELP.
        JRRELP is comprised of three loss terms: the RE loss, the KGLP loss, and the coupling loss.
        The RE loss is illustrated in the top-left quadrant, the KGLP loss is described by the top-right quadrant, and the bottom half shows the coupling loss.
        }
    \label{fig:jrrelp-overview}
    \vspace{-4ex}
\end{figure}



We propose a general framework which ties the RE and KGLP tasks cohesively into a single learning problem.
Our architecture, termed \textbf{JRRELP}---\textbf{J}ointly \textbf{R}easoning over \textbf{R}elation \textbf{E}xtraction and \textbf{L}ink \textbf{P}rediction---has the following desirable properties:
\begin{itemize}[noitemsep,topsep=0pt,label={\tiny$\blacksquare$},leftmargin=2em]
  \item \uline{Generality:}
    Our method can be applied to arbitrary RE and KGLP models to boost RE performance.
    The only assumption JRRELP makes is that both models are trained by minimizing a loss function (which is common across all successful RE and KGLP methods).
  \item \uline{Effective Information-Sharing:}
    JRRELP introduces a {\em cyclical} relationship between model parameters, enabling better information transfer between the learning tasks.
    Moreover, {\em all} parameters are shared across both the RE and KGLP tasks.
  \item \uline{Performance:}
    JRRELP boosts the performance of all baseline methods used in our evaluation.
    Additionally, JRRELP-enhanced baselines even match or improve upon the performance of more expressive RE models.
    For example, we are able to train C-GCN \citep{cgcn} to match TRE \citep{tre}, even though the latter was proposed as a stronger and significantly more expressive alternative.
  \item \uline{Efficiency:}
    JRRELP does not require any task-specific pre-training.
    It introduces a minimal overhead over the baseline methods (at most $6\%$ slower per batch).
\end{itemize}
An overview of JRRELP is shown in Figure~\ref{fig:jrrelp-overview}, and is explained in detail in Section~\ref{sec:jrrelp}.
Next, we present our proposed method and defer positioning with respect to related work until Section~\ref{sec:related_work}.

%% file: background.tex


Before presenting our method, we introduce the notation used throughout this paper, and describe the relevant learning tasks.
Let $\mathcal{D}$ describe a dataset that contains a collection of sentences.
Let $X=[x_1, x_2, \ldots x_n]$ denote a sentence, where $x_i$ represents a one-hot encoding for the $i^{\text{th}}$ sentence token (i.e., word).
Each sentence contains a subject $s = [x_{s^{\textrm{start}}}, x_{s^{\textrm{start}} + 1}, \hdots, x_{s^{\textrm{end}}}]$, that is defined as a contiguous span $(s^{\textrm{start}}, s^{\textrm{end}})$ over the sentence, and an object $o = [o_{o^{\textrm{start}}}, o_{o^{\textrm{start}} + 1}, \hdots, o_{o^{\textrm{end}}}]$, that is similarly defined.
Subjects and objects are summarized by their {\em types}, termed $s^{\textrm{type}}$ and $o^{\textrm{type}}$, respectively.
If not already given, these can be extracted by widely used parsing frameworks such as \citep{stanfordcorenlp}.
For example, consider the sentence ``\texttt{\textcolor{our_blue}{John Doe} lives in \textcolor{our_red}{Miami}}'', where the subject is shown in \textcolor{our_blue}{blue} color and the object in \textcolor{our_red}{red} color.
In this case, the subject may be tagged as having type \textcolor{our_blue}{\texttt{PERSON}} and the object may be tagged as having type \textcolor{our_red}{\texttt{CITY}}.
Several methods \citep[e.g.,][]{palstm, cgcn, aggcn} employ {\em type-substitution} during data preprocessing: substituting subjects and objects in sentences with their corresponding types.
For instance, with type-substitution our example sentence becomes
``\texttt{\textcolor{our_blue}{SUBJECT-PERSON SUBJECT-PERSON} lives in \textcolor{our_red}{OBJECT-CITY}}.''
For ease of future explanation, we assume that sentences are preprocessed using type-substitution for the remainder of this paper.
Each sentence may contain additional structural features such as part-of-speech (POS) tags, named-entity-recognition (NER) tags, and a dependency parse. Analogous to extracting entity types, these can be generated from parsing frameworks. 
We denote all such sentence features as members of a set $C$.
Finally, each sentence contains a relation, $r$, between its subject and object. 
This may either describe their lack of connection (via a special \texttt{NoRelation} token), or an existing one.
For instance, the relation between \textcolor{our_blue}{\texttt{John Doe}} and \textcolor{our_red}{\texttt{Miami}} in our example sentence would be \texttt{LivesIn}.
In summary, $\mathcal{D}$ is a set of $N$ tuples: $\mathcal{D} = \{ ( X_i, C_i, s_i, o_i, s^{\textrm{type}}_i, o^{\textrm{type}}_i, r_i ) \}_{i=1}^N$, where $N$ is the number of sentences.

\subsection{Relation Extraction}
\label{sec:re-task}
Relation extraction (RE) uses $X$, $C$, $s$, and $o$ from $\mathcal{D}$ to infer the relation $r$ between $s$ and $o$. Note that due to our type-substitution constraint, this is analogous to predicting the relation $r$ between $s^{\textrm{type}}$ and $o^{\textrm{type}}$. 
Many successful models that have been proposed to tackle this task involve learning vector embeddings for each component.
Specifically, let $N_v$, $N_r $, and $N_c$ denote the vocabulary size for sentence tokens, the number of unique relations, and the number of unique attributes in $C$, computed over the whole training dataset.
Additionally, let $D_v$, $D_r $, and $D_c$ denote the corresponding embedding sizes.
We define $\bm{V}\in\mathbb{R}^{D_v\times N_v}$, $\bm{R}\in\mathbb{R}^{D_r\times N_r}$, and $\bm{A}\in\mathbb{R}^{D_c\times N_c}$ as the vocabulary, relation, and attribute embedding matrices, respectively.
Note that $\bm{V}, \bm{R}$, and $\bm{A}$ are learnable model parameters.
Given a sentence, a subject, an object, and its attributes, their respective embedding representations are defined as:
$\bm{X} = \bm{V}X \in \smash{\mathbb{R}^{D_v \times n}}$,
$\bm{C} = \bm{A}C \in \smash{\mathbb{R}^{D_c\times c}}$,
$\bm{s^{\textrm{type}}} = \bm{V} s^{\textrm{type}} \in \smash{\mathbb{R}^{D_v}}$, and
$\bm{o^{\textrm{type}}} = \bm{V} o^{\textrm{type}} \in \smash{\mathbb{R}^{D_v}}$,
where $n$ is the number of tokens in $X$ and $c$ is the number of attributes in $C$.
Similarly, we define the embedded relation as $\bm{r} = \bm{R}r \in \mathbb{R}^{D_r}$.
Given these embeddings, most successful RE models \citep[e.g.,][]{palstm,cgcn,aggcn,tre,bert-em} can be formulated as instances of the following model:
\begin{align}
    & \bm{X} = \bm{V}X,\;
      \bm{C} = \bm{A}C,\;
      \bm{s^{\textrm{type}}} = \bm{V} s^{\textrm{type}},\;
      \bm{o^{\textrm{type}}} = \bm{V} o^{\textrm{type}},
    && \textsc{\sffamily\scriptsize\textcolor{gray}{EMBEDDING}} \\
    & \hat{\bm{r}} = f(\bm{X}, \bm{C}, \bm{s^{\textrm{type}}}, \bm{o^{\textrm{type}}}),
    && \textsc{\sffamily\scriptsize\textcolor{gray}{PREDICTION}} \\
    & p(r \mid \hat{\bm{r}}) = \textrm{Softmax}(\bm{R}\hat{\bm{r}} + \bm{b}),
    && \textsc{\sffamily\scriptsize\textcolor{gray}{PROBABILITY ESTIMATION}}
    \label{eq:re-model}
\end{align}
where $\hat{\bm{r}}$ is the inferred relation representation from a prediction model $f$.
To demonstrate how multiple RE methods fit under this formulation, we briefly describe the three baseline models used in our experiments.

\textbf{PA-LSTM.}
This model was proposed by \citet{palstm}, and centers around formulating $f$ as the combination of a one-directional long short-term memory (LSTM) network, and a custom position-aware attention mechanism.
The sentence attributes it uses are POS and NER tags, as well as SO and OO tags representing the positional offset of each token from the subject and the object respectively. The method first applies the LSTM over the concatenated sentence, POS tag, and NER tag embeddings. A relation $\hat{\bm{r}}$ is then predicted by attending the LSTM outputs with a custom position-aware attention mechanism using the SO and OO tag embeddings.

\textbf{C-GCN.}
This model was proposed by \citet{cgcn}, and formulates $f$ as a graph-convolution network (GCN) over sentence dependency parse trees. It uses the same sentence attributes as PA-LSTM, and additionally the sentence dependency parse.
Similar to PA-LSTM, the method first encodes a concatenation of the sentence, POS tag, and NER tag embeddings using a bi-directional LSTM network. The model then infers relations from these encodings by reasoning over the graph implied by a pruned version of the provided dependency tree parse.
In particular, C-GCN computes the least common ancestor (LCA) between $s$ and $o$, and uses the SO and OO tags to prune the tree around the LCA.
Afterwards, C-GCN processes the sentence encodings using a graph convolution network (GCN) defined over the pruned dependency parse tree.
The resulting representations are finally processed by a multi-layer perceptron to predict relations.

\textbf{SpanBERT.}
This model was proposed by \citet{spanbert}, and is a strong performing BERT \cite{bert}-based relation extraction method.
SpanBERT extends BERT by pre-training at the span-level.
Moreover, the model randomly masks contiguous text spans instead of individual tokens, and adds a span-boundary objective that infers masked spans from surrounding data.
In contrast to PA-LSTM and C-GCN, SpanBERT only takes into account the type-substituted sentence in its input to predict relations. $f$ is formulated as its complete architecture, with $C$ masked out.
We chose this model because it is a strong performing BERT-based RE model and it is also open-sourced, allowing to easily integrate it in our experimental evaluation pipeline.

Note that PA-LSTM, C-GCN, and SpanBERT are just three of many approaches supported by our abstract RE model formulation.
For instance, other transformer-based methods \citep{tre,bert-em,knowbert} can also be represented by using a different definition for $f$.

\subsection{Knowledge Graph Link Prediction}
\label{sec:kglp-task}

The objective in knowledge graph link prediction (KGLP) is to infer a set of objects $O$ given a question, $(s, r, ?)$, in the form of a subject-relation-object triple, missing the object.
Typically, $s$ and $o$ are nodes in a knowledge graph (KG), while $r$ represents a graph edge.
Although $\mathcal{D}$ does not necessarily provide an explicit KG to reason over, it is possible to generate one by assigning unique identifiers for all subjects, relations, and objects, For instance, these may be $s^{\textrm{type}}$ and $o^{\textrm{type}}$ for subjects and objects respectively, and the relation itself.
Although we assume that these identifiers are used 
(as they are available in our training data $\mathcal{D}^{\textrm{train}}\subset \mathcal{D}$), we emphasize that our method is not limited to datasets with these characteristics. Instead our framework supports any $\mathcal{D}$ that specifies a mapping to a pre-existing KG, or where it is possible to define other unique identifiers. This is a very weak constraint.
Therefore, given a sentence with $s$, $o$, and $r$, we can use the subject and object types---$s^{\textrm{type}}$ and $o^{\textrm{type}}$, respectively---to form a KG whose edges are represented by each $r$ and nodes by each $s^{\textrm{type}}$ and $o^{\textrm{type}}$.
For ease of notation, we assume that each term is a one-hot encoding of the corresponding identifier.

Due to the {\em type-substitution} preprocessing step described in Section~\ref{sec:background}, all types are included in the sentence token vocabulary.
Thus, we obtain KG component embeddings by:
$\bm{s^{\textrm{type}}} = \bm{V} s^{\textrm{type}} \in \smash{\mathbb{R}^{D_v}}$,
$\bm{o^{\textrm{type}}} = \bm{V} o^{\textrm{type}} \in \smash{\mathbb{R}^{D_v}}$, and
$\bm{r} = \bm{R} r \in \smash{\mathbb{R}^{D_r}}$.
Multiple existing KGLP methods can be characterized in terms of the following abstract model:
\begin{align}
    & \bm{s^{\textrm{type}}} = \bm{V} s^{\textrm{type}},\;
      \bm{r} = \bm{R} r,
    && \textsc{\sffamily\scriptsize\textcolor{gray}{EMBEDDING}} \\
    & \bm{z} = g(\bm{s^{\textrm{type}}}, \bm{r}),
    && \textsc{\sffamily\scriptsize\textcolor{gray}{MERGE}} \\
    & p(O \mid o^{\textrm{type}}, \bm{z}) = \textrm{Sigmoid}(\bm{V}_{o^{\textrm{type}}}\bm{z} + \bm{b}),
    && \textsc{\sffamily\scriptsize\textcolor{gray}{PROBABILITY ESTIMATION}}
    \label{eq:kglp-model}
\end{align}
where $\bm{z}$ is a merged representation of $\bm{s^{\textrm{type}}}$ and $\bm{r}$.
Note that the set of available object embeddings $\bm{V}_{o^{\textrm{type}}} \subset \bm{V}$ contains {\em only} valid (in the type-checking sense) object embeddings.
Previous work \citep{coper} shows that multiple KGLP methods fit under this formulation. 
While certain early KGLP methods \citep{bordes2013translating, yang2015embedding, transr_ctranr, transd, trouillon2016complex} do not fit under this formulation, we note that they may be accommodated by a simple reconfiguration of Equation~\ref{eq:kglp-model} to their respective scoring terms.
We now provide the definition of ConvE \citep{dettmers2018conve} under this formulation, because we use ConvE as our KGLP model in our experiments.
While we acknowledge that ConvE is not the current state-of-the-art (SoTA) KGLP approach, it performs very well while using only a fraction of the parameters current SoTA \cite{coper, GAAT} methods require, thus making it more efficient.
Moreover, ConvE is an example of a KGLP method which cannot be restructured to infer $r$ from $s$ and $o$, making it infeasible to use with any of the previous joint RE and KGLP frameworks \citep[e.g.,][]{lfds,weston-2013}.
Note that, our results can only be further enhanced by using a stronger KGLP approach and thus this choice should not affect our conclusions.

\textbf{ConvE.}
ConvE is defined by using the following merge function in our abstract model formulation:
\begin{align}
    & g(\bm{s^{\textrm{type}}}, \bm{r}) = \text{Conv2D}(\text{Reshape}([\bm{s^{\textrm{type}}}; \bm{r}]),
    && \textsc{\sffamily\scriptsize\textcolor{gray}{MERGE}}
\end{align}
where ``Conv2D'' is a 2D convolution operation and ``$\text{Reshape}([\bm{s^{\textrm{type}}}; \bm{r}])$'' first concatenates $\bm{s^{\textrm{type}}}$ and $\bm{r}$ and then reshapes the resulting vector to be a square matrix, so that a convolution operation can be applied to it.

%% file: method.tex
\label{sec:jrrelp}

As mentioned in Section~\ref{sec:intro}, the RE and KGLP tasks are tightly coupled.
Given a sentence $X$ (e.g., ``\texttt{\textcolor{our_blue}{Miami} is in \textcolor{our_red}{Florida}}'') that contains a subject $s$ (e.g., \texttt{\textcolor{our_blue}{Miami}}) and an object $o$ (e.g., \texttt{\textcolor{our_red}{Florida}}), the goal of RE is to predict the relation $r$ (e.g., \texttt{locatedIn}), between $s$ and $o$, that the sentence describes. Similarly, the goal of KGLP is to infer a set of objects $O$ using $r$ and $s$, such that the inferred objects correspond to correct subject-relation-object triples, and where $o\in O$ (this is known because the sentence $X$ describes this relationship).
Based on this observation, we propose JRRELP, a multi-task learning framework that explicitly accounts for this relationship between RE and KGLP.
JRRELP trains a RE model, $p_{\textrm{\tiny RE}}$, that is defined using our abstract formulation from Section~\ref{sec:re-task} and a KGLP model, $p_{\textrm{\tiny KGLP}}$, that is defined using our abstract formulation from Section~\ref{sec:kglp-task}, jointly, using four key ideas:
\begin{enumerate}[noitemsep,topsep=0pt,leftmargin=2em]
    \item \uline{Parameter Sharing:}
        $p_{\textrm{\tiny RE}}$ and $p_{\textrm{\tiny KGLP}}$ share all of the embedding parameters.
        This corresponds to the matrices $\bm{V}$, $\bm{R}$, and $\bm{A}$ from Sections~\ref{sec:re-task}~and~\ref{sec:kglp-task}. Moreover, all parameters between RE and KGLP methods are also shared.
    \item \uline{Joint Training:}
        The two models are trained jointly by optimizing a single objective function.
        This function contains terms that correspond to the RE objective function, the KGLP objective function, as well as a prediction coupling loss term.
    \item \uline{Cyclical Coupling:}
        Our joint loss terms establish a cyclical relationship between the embedding parameters, that tightly couples the RE and KGLP tasks.
        This is because the RE model uses $\bm{V}$ (which includes $\bm{V}_{o^{\textrm{type}}}$) to predict relation representations that are then compared to $\bm{R}$ to produce distribution over relations. Reciprocally, the KGLP model uses $\bm{R}$ to generate object embeddings that are compared to $\bm{V}_{o^{\textrm{type}}}$ to produce distributions over objects.
    \item \uline{Unmodified Evaluation:} JRRELP does not introduce any additional terms when evaluating $p_{\textrm{RE}}$. Thus, rather than enhancing $p_{\textrm{RE}}$ by increasing its capacity, JRRELP does this by altering its training trajectory.
\end{enumerate}
We now provide details on how each term of the joint training objective function is defined.

\textbf{RE Loss.}
The first term corresponds to the standard loss function used to train the RE model.
This loss function is defined as follows (where we use the notation introduced in Section~\ref{sec:re-task}):
\begin{equation}
    \mathcal{L}_{\textrm{\tiny RE}} = \sum_{i=1}^N \textrm{SCE}(r_i, p_{\textrm{\tiny RE}}(r_i \mid X_i, C_i, s^{\textrm{type}}_i, o^{\textrm{type}}_i)),
\end{equation}
where ``SCE'' represents the softmax cross-entropy loss function, and $p_{\textrm{\tiny RE}}$ is defined as in Equation~\ref{eq:re-model}:
\begin{equation}
    p_{\textrm{\tiny RE}}(r_i \mid X_i, C_i, s^{\textrm{type}}_i, o^{\textrm{type}}_i) = \textrm{Softmax}(\bm{R} f_{\textrm{\tiny RE}}(\bm{X}_i, \bm{C}_i, \bm{s^{\textrm{type}}}_i, \bm{o^{\textrm{type}}}_i) + \bm{b}_{\textrm{\tiny RE}}),
\end{equation}
where $f_{\textrm{\tiny RE}}$ is the specific prediction function used by our RE model.
Although this loss term assumes that a {\em single} relation exists between a subject and an object in a sentence, it is consistent with the loss term utilized by our baselines and is also appropriate for our widely used benchmark datasets described in Section \ref{sec:experiments}.
Additionally, we note that this does not restrict the applicability of JRRELP to single-relation extraction problems.
For instance, ``SCE'' can be substituted for binary-cross entropy (BCE) in the case of having multiple applicable relations.

\textbf{KGLP Loss.}
The second term corresponds to a popular loss function which is often used to train KGLP models.
This loss function is defined as follows (where we use the notation introduced in Section~\ref{sec:kglp-task}):
\begin{equation}
    \mathcal{L}_{\textrm{\tiny KGLP}} = \sum_{i=1}^N \textrm{BCE}(O_i, p_{\textrm{\tiny KGLP}}(O_i \mid s^{\textrm{type}}_i, o^{\textrm{type}}_i, r_i)),
\end{equation}
where 
$p_{\textrm{\tiny KGLP}}$ is defined as in Equation~\ref{eq:kglp-model}:
\begin{equation}
    p_{\textrm{\tiny KGLP}}(O_i \mid s^{\textrm{type}}_i, o^{\textrm{type}}_i, r_i)) = \textrm{Sigmoid}(\bm{V}_{o^{\textrm{type}}_i} g_{\textrm{\tiny KGLP}}(\bm{s^{\textrm{type}}_i}, \bm{r}_i) + \bm{b}_{\textrm{\tiny KGLP}}),
\end{equation}
where $g_{\textrm{\tiny KGLP}}$ is the specific merge function used by our KGLP model.
Note here that $O_i$ is a set of objects that can be constructed automatically given all of the training data and conditioned on $s^{\textrm{type}}_i$ and $r_i$, as described in Section~\ref{sec:kglp-task}.
We also acknowledge that certain KGLP methods \citep{bordes2013translating, yang2015embedding, transr_ctranr, transd, trouillon2016complex} cannot be represented by this loss term.
However, this does not detract from the generality of the proposed framework because they can be accommodated by changing this term to their respective objective functions.

\textbf{Coupling Loss.}
The third term penalizes inconsistencies between the predictions of the RE and KGLP models.
It is defined as follows:
\begin{equation}
    \mathcal{L}_{\textrm{\tiny COUPLING}} = \sum_{i=1}^N \textrm{BCE}(O_i, p_{\textrm{\tiny COUPLING}}(O_i \mid X_i, C_i, s_i, o_i, s^{\textrm{type}}_i, o^{\textrm{type}}_i)),
\end{equation}
where:
\begin{equation}
    p_{\textrm{\tiny COUPLING}}(O_i \mid \hdots) =
    \textrm{Sigmoid}(\bm{V}_{o^{\textrm{type}}_i} g_{\textrm{\tiny KGLP}}(\bm{s^{\textrm{type}}_i}, \textcolor{our_red}{f_{\textrm{\tiny RE}}(\bm{X}_i, \bm{C}_i, \bm{s^{\textrm{type}}}_i, \bm{o^{\textrm{type}}}_i}) + \bm{b}_{\textrm{\tiny KGLP}}),
\end{equation}
where we have omitted the conditioning variables for brevity.
The key difference between this loss term and the KGLP loss term is shown in \textcolor{our_red}{red} color.
Specifically, the relations embeddings --- $\bm{r}_i$ --- computed by $r_i$ in the KGLP loss term, are replaced by the predicted relation embeddings $\hat{\mathbf{r}}_i$ from $f_{\textrm{RE}}$.
This term aligns the RE and KGLP methods by making the first compatible with the second, and enhances the overall performance of our framework.

\subsection{JRRELP Objective Function}

The JRRELP objective function is formed by putting together the above three terms:
\begin{equation}
    \mathcal{L}_{\textrm{\tiny JRRELP}} = \mathcal{L}_{\textrm{\tiny RE}} + \lambda_{\textrm{\tiny KGLP}} \mathcal{L}_{\textrm{\tiny KGLP}} + \lambda_{\textrm{\tiny COUPLING}} \mathcal{L}_{\textrm{\tiny COUPLING}},\label{eq:jrrelp}
\end{equation}
where $\lambda_{\textrm{\tiny KGLP}} \geq 0$ and $\lambda_{\textrm{\tiny COUPLING}} \geq 0$ are model hyperparameters that need to be tuned properly.
We note that, while in principle $\lambda_{\textrm{\tiny KGLP}}$ and $\lambda_{\textrm{\tiny COUPLING}}$ can vary independently, in our experiments we set both to the same value for simplicity and cheaper hyperparameter tuning. Furthermore, we observed no negative impact in performance.

Most importantly, due to the JRRELP parameter sharing and the use of this loss function, our framework introduces a cyclical relationship between the RE and KGLP models that couples them together very tightly.
Specifically, the RE model predicts relation embeddings using $\bm{V}$ that it compares to $\bm{R}$ to produce distributions over relations.
The KGLP model on the other hand predicts object embeddings using $\bm{R}$ that it compares to $\bm{V}$ to produce distributions over objects.
It is mainly this cyclical relationship along with the coupling loss term that result in both the RE and KGLP models benefiting from each other and serves to enhance the performance and robustness of RE methods.
An overview of JRRELP is shown in Figure~\ref{fig:jrrelp-overview}.

Note that, even though JRRELP minimizes the joint three-task objective function shown in Equation~\ref{eq:jrrelp}, at test time we only use the RE model to predict relations between subjects and objects.
Thus, JRRELP can be thought of as a framework which alters the learning trajectory of an RE model, rather than increase its capacity through using additional model parameters.

%% file: experiments.tex
We empirically evaluate the performance of JRRELP over two existing relation extraction baselines on two widely used supervised benchmark datasets.
Our primary objective is to measure the importance of a joint RE and KGLP objective in environments where learning over both tasks is restricted {\em only} to data available in a relation extraction dataset.
This serves to simulate how effective JRRELP may be in real-world applications where a pre-existing KG is not available for a given RE task.
Additionally, we perform an ablation study to examine the impact each part of JRRELP has on its overall performance.

\textbf{Datasets.}
We use the TACRED \cite{palstm} and SemEval 2010 Task 8 \cite{semeval} datasets for our experiments, which are commonly used in prior literature \citep[e.g.,][]{palstm, cgcn, aggcn, bert-em}. 
Table \ref{tab:dataset_statistics} shows their summary statistics.
As mentioned in Section~\ref{sec:background}, for both datasets we utilize the following sentence attributes: NER tags, POS tags, subject/object offsets, and dependency tree structure.
For the KGLP task in JRRELP, we construct the KG by generating ($s^{\textrm{type}}$, $r$, $o^{\textrm{type}}$) triples automatically, for each training sentence.
We then ask questions of the form $(s^{\textrm{type}}, r, ?)$, where the answer belongs to a set of applicable objects $O$.

\textbf{Setup.}
We perform our experiments on TACRED consistent with prior literature \citep{palstm, cgcn, aggcn}.
We use the same type-substitution policy where we replace each subject and object in a sentence with their corresponding NER types.
Additionally, we evaluate our models using their micro-averaged $\texttt{F1}$ scores.
Finally. we report the test metrics of the model with the best validation $\texttt{F1}$ score over five independent runs.
While SemEval 2010 Task 8 is traditionally evaluated without type-substitution, \citet{cgcn} point out that this causes models to overfit to specific entities, and does not test their ability to generalize to unseen data.
They address this by masking these entities using their types.
Therefore, to examine JRRELP's generalization capabilities, we perform the same type-substitution procedure, and evaluate on the transformed dataset (denoted as SemEval-MM).
Consistent with prior work  \citep{palstm,cgcn,aggcn,tre, bert-em}, we report the macro-averaged \texttt{F1} scores. 
Because SemEval(-MM) does not contain a validation set, we subsample $800$ examples from the training set to use as a validation set.
\input{tables/dataset_statistics}

\textbf{Models.}
We illustrate the generality of JRRELP by evaluating it on baselines from both classes of RE approaches:\footnote{Refer to Section~\ref{sec:related_work} for their definitions.}
Two sequence-based models (PA-LSTM and SpanBERT), and a graph-based model (C-GCN). 
We join all three baselines with the KGLP method ConvE.
We distinguish between our baselines and their JRRELP variants by boxing their model names (e.g. \framebox{PA-LSTM} is the JRRELP extended version of PA-LSTM). All models can be found in our repository: https://github.com/gstoica27/JRRELP.git.
\textbf{Results.}
\input{tables/model_results}
We report our overall performance results on TACRED in Table \ref{tab:results}. We observe that JRRELP consistently outperforms it's baseline variants over their \texttt{F1} and precision metrics.
In particular, we find that JRRELP improves all baseline model performances by at least $.6\%$ \texttt{F1}, and yields improvements of up to $4.1\%$ in precision. 
Furthermore, JRRELP bridges the performance gap between several methods, {\em without} altering their model capacities. 
Notably, JRRELP extended PA-LSTM matches the reported C-GCN performance, whose JRRELP variant matches TRE \cite{tre} --- a significantly more expressive transformer-based approach. 
These results suggest that the true performance ceiling of reported relation extraction approaches may be significantly higher than their reported results, and that JRRELP serves as a conduit towards achieving these performances.
Results on SemEval-MM indicate a similar pattern to TACRED: JRRELP improves performance across all baselines. 
This illustrates the effectiveness of JRRELP's framework in environments with little data.  


\input{tables/ablation_tests}

\textbf{Ablation Experiments.}
To examine the effects of JRRELP's $\mathcal{L}_\text{KGLP}$ and $\mathcal{L}_{\text{COUPLING}}$ over the traditional relation extraction objective, $\mathcal{L}_{\text{RE}}$, we perform an ablation study with each term removed on methods from both RE approach classes: sequence-based (PALSTM) and graph-based (C-GCN). Table \ref{tab:ablation_tests} shows the \texttt{F1} results. Metrics for each dataset are reported in the same manner as previous results. 
All ablation performances illustrate the importance of $\mathcal{L}_\text{KGLP}$ and $\mathcal{L}_{\text{COUPLING}}$ as part of JRRELP's framework, as their respective models are worse than the full JRRELP architecture: they exhibit performance drops up to $.8\%$ \texttt{F1} respectively. Moreover, we observe the largest performance drop from the removal of $\mathcal{L}_\text{COUPLING}$ -- which removes JRRELP's consistency constraint between RE and KGLP models. This highlights importance of establishing this relationship while training to achieve strong performance.


%% file: tables/dataset_statistics.tex
\begin{table*}[ht]
	\begin{center}
		\small
		\caption{Dataset statistics. Here,  \# Train, \# Validation, and \# Test denote the number of questions used for training, validation, and testing. \# Relations describes the number of distinct relation in each dataset, Avg. Tokens refers to the average number of tokens in each dataset sentence,
	and \% Negatives indicates the percentage of data where there is "no relation" between subjects and objects. \\}
	\label{tab:dataset_statistics}
		\begin{tabular}{ccccccc} 
        \toprule
		\rule{0pt}{9pt} 
		\textbf{Dataset} & \# Train &  \# Validation & \# Test & \# Relations & Avg. Tokens & \% Negatives \\ 
        \midrule
		TACRED              &   68,124 &    22,631 &  15,509   & 42 & 36.4 & 79.5\% \\
		SemEval-MM          &   8,000  & -         &  2,717    & 19 & 19.1 & 17.4\% \\
        \bottomrule
		\end{tabular}
	\end{center}
	
\end{table*}

%% file: tables/model_results.tex
\begin{table*}[!t]
\centering
\caption{Results reported by our own experiments are marked by $^*$. The remainder are taken from \cite{tre} and \cite{knowbert}.
All numbers are expressed as percentages.
$\dagger$ denotes experiments performed using additional data other than provided by the respective models.
``--'' denotes missing results from the respective publications. 
``SemEval-MM'' denotes the Masked-Mention version of the SemEval dataset.
\\}
\begin{adjustbox}{width=\textwidth}
\begin{tabular}{llccccccccc}
\toprule
\multirow{2}{*}{\textbf{Dataset}} & \multirow{2}{*}{\textbf{Metric}} & \multicolumn{9}{c}{\textbf{Models}} \\
\cmidrule{3-11}
& & \textbf{C-AGGCN} & \textbf{TRE} & $\textbf{BERT}_{EM}$ & \textbf{PA-LSTM} & \framebox{\textbf{PA-LSTM}} &  \textbf{C-GCN}  & \framebox{\textbf{C-GCN}} & \textbf{SpanBERT} & \framebox{\textbf{SpanBERT}} \\
\midrule                
\multirow{3}{*}{TACRED}
  & \texttt{Precision}   & 73.1 & 70.1 & --                      & 65.7  &  67.8$^*$ & 69.9    & \textbf{74.1}$^*$  & 69.2* & 74.0* \\
  & \texttt{Recall}  & 64.2 & 65.0     & --                      & 64.5  &  65.0$^*$ & 63.3    &  61.9$^*$          & 71.2* & 67.3* \\
  & \texttt{F1}      & 68.2 & 67.4     & \textbf{71.5}$^\dagger$ & 65.1  &  66.4$^*$ & 66.4    &  67.4$^*$          & 70.2* & 70.8* \\
  \midrule
 \multirow{3}{*}{SemEval-MM} 
  & \texttt{Precision} & -- & -- & -- & 75.2 & 74.8 & 76.5 & 76.9 & 81.2 & 82.7 \\
  & \texttt{Recall} & -- & -- & -- & 78.0 & 80.6 & 79.5 & 80.3 & 86.1 & 85.2 \\
  & \texttt{F1} & -- & -- & -- & 76.6 & 77.6 & 78.0 & 78.5 & 83.6 & 83.9 \\
\bottomrule
\end{tabular}
\end{adjustbox}
\label{tab:results}
\vspace{-1ex}
\end{table*}

%% file: tables/ablation_tests.tex
\begin{table*}[ht]
\centering
\tiny
\caption{TACRED \texttt{F1} results from our ablation study. $\dagger$ denotes experiments conducted without $\mathcal{L}_\text{COUPLING}$, and $\ddagger$ marks those run without $\mathcal{L}_{\text{KGLP}}$. \\}
\begin{adjustbox}{width=\textwidth}
\begin{tabular}{llcccc|cccc}
\toprule
\multirow{2}{*}{\textbf{Dataset}} & \multirow{2}{*}{\textbf{Metric}} & \multicolumn{8}{c}{\textbf{Ablation Experiments}} \\
\cmidrule{3-10}
& &  \textbf{PALSTM} & \textbf{\framebox{PA-LSTM}} & \textbf{\framebox{PALSTM}$^	\dagger$} & \textbf{\framebox{PALSTM}$^	\ddagger$} & \textbf{C-GCN} & \textbf{\framebox{C-GCN}} & \textbf{\framebox{C-GCN}$^\dagger$} & \textbf{\framebox{C-CGCN}$^\ddagger$}\\
\midrule
TACRED & \texttt{F1} & 65.1 & 66.4 & 65.6 & 66.3 & 66.4 & 67.4 & 66.8 & 67.0  \\
SemEval-MM & \texttt{F1} & 76.6 & 77.6 & 76.8 & 77.3 & 78.0 & 78.5 & 78.1 & 78.4 \\

\bottomrule
\end{tabular}
\end{adjustbox}
\label{tab:ablation_tests}
\end{table*}

%% file: related_work.tex

There are three areas of research that are related to the method we propose in this paper.
In this section, we discuss related work in each area and position JRRELP appropriately.

\textbf{Relation Extraction.}
Existing RE approaches can be classified in two categories: sequence-based, and graph-based methods.
Given a sentence in the form of a sequence of tokens, sequence-based models infer relations by applying recurrent neural networks \cite{zhou-etal-2016-attention, palstm}, convolutional neural networks \cite{zeng-etal-2014-relation, nguyen-grishman-2015-relation, wang-etal-2016-relation}, or transformers \cite{tre, bert-em, spanbert, knowbert}.
In addition to the sentence,
graph-based methods use the structural characteristics of the sentence dependency tree to achieve strong performance. \cite{peng2017} apply an n-ary Tree-LSTM \cite{treelstm} over a split dependency tree, while \cite{cgcn, aggcn} employ a graph-convolution network (GCN) over the dependency tree.

\textbf{Knowledge Graph Link Prediction.}
Existing KGLP approaches broadly fall under two model classes: single-hop and multi-hop.
Given a subject and a relation, single-hop models infer a set of objects by mapping the subject and relation respectively to unique learnable finite dimensional vectors (embeddings) and jointly transforming them to produce an object set. These approaches can be translational \cite{bordes2013translating} over the embeddings, multiplicative \cite{yang2015embedding, trouillon2016complex}, or a combination of the two
\cite{dettmers2018conve, transr_ctranr, transd, tucker, coper, GAAT}.
On the other hand, multi-hop approaches determine object sets by finding paths in the KG connecting subjects to the objects, and primarily consist of path-ranking methods 
\cite{lao2011random, gardner2013improving, neelakantancompositional, guutraversing, toutanova2016compositional, minerva, salesforce}.

\textbf{Joint Frameworks.}
Several approaches  \cite{weston-2013, han, lfds, long_tail, bag_re_kglp} have explored using the additional supervision provided by a KG to benefit relation extraction model performance. Of these, we believe \cite{weston-2013, han, lfds} are most similar to our work. \cite{weston-2013} proposes a framework which utilizes a KGLP model, TransE \cite{bordes2013translating}, as an additional re-ranking term when evaluating an RE model. While employing TransE as a re-ranker improves performance, their framework trains TransE and the respective RE approach separately without parameter sharing. This only allows very restricted information sharing during evaluation.
\cite{han} proposes a dual-attention framework for jointly learning KGLP and RE tasks by computing a weight distribution over training data and shares parameters between tasks. However, like \cite{weston-2013}, \cite{han} limits KGLP model selection to those which can reformulated as inferring relations from subjects and objects. This excludes a large number of recent methods \cite{dettmers2018conve, tucker, minerva, salesforce, coper, GAAT} which cannot be reframed in this way. \cite{lfds} also presents a joint framework, LFDS, for training relation extraction approaches via KGLP objectives. In particular, the architecture introduces a similar objective to $\mathcal{L}_{\text{COUPLING}}$, but can only support the same class of KGLP methods as in \cite{weston-2013, han}. Moreover, LFDS requires KGLP pre-training, and does not share core parameters such as relation representations between RE and KGLP methods. This can create domain-shift between the two respective models and impact performance.

JRRELP improves upon previous literature by providing a single joint objective which simultaneously addresses all their aforementioned limitations. First, JRRELP proposes an abstract framework which supports many RE and KGLP methods through three standard-based loss terms. Second, JRRELP shares all its parameters between KGLP and RE tasks, and establishes a novel cyclical learning structure over core parameters. Third, RE and KGLP tasks are jointly trained without any problem-specific pretraining required, enabling tasks to benefit from each other simultaneously during training. Fourth, JRRELP's structure facilitates suport for RE and KGLP methods with minimal implementation changes: only requiring their respective substitutions into $f$ and $g$.




%% file: conclusion.tex
We propose JRRELP, a novel framework that improves upon existing relation extraction approaches by leveraging insights from the complementary problem of knowledge graph link prediction. JRRELP bridges these two tasks through an abstract multi-task learning framework that jointly learns RE and KGLP problems by unconstrained parameter sharing. We exhibit this generality be extending three diverse relation extraction methods, and improve their performances. 